# Use of "off-the-shelf" information extraction algorithms in clinical informatics: a feasibility study of MetaMap annotation of Italian medical notes

Emma Chiaramello[a], Francesco Pinciroli[a,b], Alberico Bonalumi[c], Angelo Caroli[c], , Gabriella Tognola[a]

[a] Istituto di Elettronica e di Ingegneria dell'Informazione e delle Telecomunicazioni (IEIIT), Consiglio Nazionale delle Ricerche (CNR), Piazza L. da Vinci, 32, 20133 Milano, Italy.

[b] e-HealthLAB, Dipartimento di Elettronica, Informazione e Bioingegneria (DEIB), Politecnico di Milano, Piazza L. da Vinci, 32, 20133 Milan, Italy.

[c] UOC Sistemi Informativi e Informatici, Fondazione IRCCS Ca' Granda Ospedale Maggiore Policlinico Milano, Via Francesco Sforza, 35, 20122 Milano, Italy.

**Corresponding author:**

Gabriella Tognola

Istituto di Elettronica e di Ingegneria dell'Informazione e delle Telecomunicazioni (IEIIT), Consiglio Nazionale delle Ricerche (CNR), Piazza L. da Vinci, 32, 20133 Milano, Italy.
gabriella.tognola@ieiit.cnr.it

**Abstract**

Information extraction from narrative clinical notes is useful for patient care, as well as for secondary use of medical data, for research or clinical purposes. Many studies focused on information extraction from English clinical texts, but less dealt with clinical notes in languages other than English.

This study tested the feasibility of using "off the shelf" information extraction algorithms to identify medical concepts from Italian clinical notes. Among all the available and well-established information extraction algorithms, we used MetaMap to map medical concepts to the Unified Medical Language System (UMLS). The study addressed two questions: (Q1) to understand if it would be possible to properly map medical terms found in clinical notes and related to the semantic group of "Disorders" to the Italian UMLS resources; (Q2) to investigate if it would be feasible to use MetaMap as it is to extract these medical concepts from Italian clinical notes.

We performed three experiments: in EXP1, we investigated how many medical concepts of the "disorders" semantic group found in a set of clinical notes written in Italian could be mapped to the UMLS Italian medical sources; in EXP2 we assessed how the different processing steps used by MetaMap, which are English dependent, could be used in Italian texts to map the original clinical notes on the Italian UMLS sources; in EXP3 we automatically translated the clinical notes from Italian to English using Google Translator, and then we used MetaMap to map the translated texts.

Results in EXP1 showed that the Italian UMLS Metathesaurus sources covered 91% of the medical terms of the "Disorders" semantic group, as found in the studied dataset. We observed that even if MetaMap was built to analyze texts written in English, most of its processing steps worked properly also with texts written in Italian. MetaMap identified correctly about half of the concepts in the Italian clinical notes. Using MetaMap's annotation on Italian clinical notes instead of a simple text search improved our results of about 15 percentage points. MetaMap's annotation of Italian clinical notes showed recall, precision and F-measure equal to 0.53, 0.98 and 0.69, respectively. Most of the failures were due to the impossibility for MetaMap to generate meaningful variants for the Italian language, suggesting that modifying MetaMap to allow generating Italian variants could improve the performance. MetaMap's performance in annotating automatically translated English clinical notes was in line with findings in the literature, with similar recall (0.75), F-measure (0.83) and even higher precision (0.95). Most of the failures were due to a bad Italian to English translation of medical terms, suggesting that using an automatic translation tool specialized in translating medical concepts might be useful to obtain better performances.

In conclusion, performances obtained using MetaMap on the fully automatic translation of the Italian text are good enough to allow to use MetaMap "as it is" in clinical practice.

# 1. Introduction

In the last few years, the development of tools to extract information from clinical digital documents has become one of the most interesting research areas in the field of medical information technology, as patient information has the potential to make a significant impact in many aspects of healthcare and biomedical research [1]. Recent works have demonstrated this potential for individual patient care (such as with clinical decision support systems [2,3]), to improve health literacy [4], for public health (e.g. in bio-surveillance [5]), and in biomedical research (e.g. in cohort identification [6,7], in the discovery of unknown correlations between diseases [8,9] or in pharmacovigilance [10,11]).

Although portion of patient information exists as structured data, a huge part of clinical data is still in the narrative form. Unstructured texts allow clinicians to easily report on various types of information about patients, i.e. symptoms, pathologies, medical histories, findings, medical treatments and drugs, and to make hypotheses, diagnoses, prescriptions, and suggestions using a free-form structure. Clinical notes can be very heterogeneous, with flexible formatting, short and telegraphic phrases and typically contain abbreviations, acronyms, misspellings and terms specific to the particular medical context [1,12]. This narrative text is adopted in many different clinical settings, e.g. in ambulatory clinical practice, and presents several difficulties to be analyzed for automated information extraction and for secondary use for research or clinical purposes.

The first task in automated information extraction from clinical notes is semantic annotation of relevant concepts in the text, i.e. the detection of noun phrases and their classification according to the semantic categories of interest [13,14]. Semantic annotation requires two components: a linguistic processing tool to detect noun phrases and a knowledge source to classify the identified concepts in the proper semantic category. The goodness of the annotation depends both on the efficiency of the linguistic tool and the quality and coverage of the knowledge sources [15,16].

Natural Language Processing (NLP) methods are those linguistic processing tools that are needed to detect noun phrases from unstructured text. Many NLP tools were originally developed to extract information from biomedical literature. In the last few years these tools have been largely adapted and used in the analysis of clinical notes [1,12]. Among the earliest attempts to develop NLP applications for the medical domain, the Linguistic String Project-Medical Language Processor (LSP-MLP) developed in 1986 [17], and the Medical Language Extraction and Encoding System (MedLEE) [18,19] were the most known examples. A more recent NLP tool by the National Library of Medicine (NLM) – MetaMap [20] – was developed to extract medical concepts from biomedical literature. In the last decade MetaMap has been also used to analyze narrative texts different from scientific papers, such as e-mails [21], patients-reports [22] and clinical notes (e.g. to identify specific respiratory conditions [5], epilepsy [6], to extract clinical problems for inclusion in the patient's electronic problem list [23,24], to discover adverse drug reactions [25]). MetaMap uses computational linguistic techniques to identify words and phrases in English free text and map them to the Unified Medical Language System (UMLS), a knowledge source developed by the US National Library of Medicine as a starting point for the development of biomedical language processing algorithms [26].

Recently, efforts were made to develop new and *ad hoc* information extraction algorithms based on linguistic processing tools for the identification of medical concepts in biomedical texts written in languages different from English, such as for the Spanish [27], French [28,29], Portuguese [30] and Swedish [14] languages. Only a few studies focused on using already existing, or "off the shelf", linguistic algorithms to process non-English texts [15,31,32].

The goal of the present study was to understand if using a linguistic tool developed to process English text, such as MetaMap, could be suitable to extract medical concepts from clinical texts written in other languages, in particular in Italian. In the view to assess the ease of use of an annotation tool in clinical practice, our idea was not to develop a new "Italian MetaMap", but to slightly adapt the existing tool. To the best of our knowledge, no previous works focused on the identification of medical concepts from Italian clinical notes using MetaMap.

Dealing with Italian language is a challenging task, as Italian language is more inflective than English and is characterized by a higher number of compounds of words, particularly in the medical lexicon. In addition, the coverage of medical terminology resources in Italian is lower than in English.

The present study addressed two specific research questions:

(Q1)  are the medical terms found in Italian clinical notes and related to the semantic group of "Disorders" properly mapped in the Italian medical terminological resources?

(Q2)  is it feasible to use MetaMap "as it is" to identify medical concepts related to the semantic group of "Disorders" from Italian clinical notes?

To answer to Q1, we carried out experiment "EXP1" in which we investigated how many medical concepts in a set of unstructured clinical notes written in Italian could be mapped to the UMLS Italian medical sources.

To answer to Q2, we followed two strategies. The first one was to test the feasibility of using MetaMap to directly analyze Italian text. The first strategy was tested in experiment "EXP2" in which (i) we first assessed how the different processing steps used by MetaMap, which are English dependent, could be used for other languages, such as Italian; and then (ii) we run MetaMap to map the original Italian texts using a modified version of the UMLS knowledge source consisting in the Italian UMLS Metathesaurus dictionaries only.

In the second strategy we followed to address question Q2, we assessed the feasibility of using MetaMap to analyze an English translated version of the original Italian text. We carried the experiment "EXP3" out, in which we automatically translated the clinical notes from Italian to English using Google Translator, and then we used MetaMap with its default knowledge source to annotate the translated texts. The approach of combining Google Translator and MetaMap seemed feasible to be applied in the context of extracting medical concepts in a clinical setting for several reasons: it is based on open source tools, it has low computational impact so that it can be used in clinical practice, and it does not require high expertise or training skills. Last but not least, the combination of Google Translator and MetaMap to identify medical concepts in texts written in a language different from English was described earlier in [31,32] to be a promising method. It is to note that our study differed from [31,32] because we tested the approach on a different language (Italian in our study, Spanish in [31,32]) and on a different type of text (we used clinical texts while in [31,32] they analyzed articles from biomedical literature), so it would be interesting to compare if our findings would be different from those reported by the Spanish researchers as a result of the different language and the different type of document.

## 2. Materials and methods

### 2.1. MetaMap processing steps

MetaMap is based on six processing steps: (1) tokenization, (2) parsing, (3) variant generation, (4) candidate retrieval, (5) candidate evaluation and (6) mapping construction.

As shown in Table 1, some processing steps are based on rules and lexical resources built on English grammar and are therefore strictly language specific, while others use algorithms that are not based on English grammar and they can therefore be used also with languages different from English. Step (1), tokenization, uses a set of rules built on English language for the identifications of tokens, sentence boundaries and acronyms/abbreviations. Although the rules (e.g. the characters of punctuation that identify the sentence boundaries, or the white space that separates different words) were optimized to tokenize English texts, they can be used to tokenize texts in other languages based on the Latin writing system, thus tokenization works properly even when analyzing Italian texts. Steps (2) and (3), i.e. parsing and variant generation, depend on the SPECIALIST lexicon, the SPECIALIST minimal commitment parser, and the Lexical Variant Generation (LVG), respectively. As shown in Table 2, all these resources are strictly built on English language; thus, steps (2) and (3) do not work properly in analyzing Italian texts. Step (4), i.e. candidate retrieval, depends uniquely on the language of the Metathesaurus knowledge source used for the analysis, thus it works properly both on English and other languages for which it is possible to define a Metathesaurus-like knowledge source, such as Italian. Finally, steps (5) and (6), i.e. candidate evaluation and mapping construction, are completely language independent, thus they work properly with texts written in languages different from English, such as Italian.

Table 1. MetaMap's processing steps.

| Processing step | Goal | Linguistic resources involved | How does it work | Language dependent | Feasibility in analyzing Italian text |
|---|---|---|---|---|---|
| 1. Tokenization | Identification of tokens, sentence boundaries and acronyms/abbreviations (AA) | - | The tokenizer uses a set of rules to identify tokens, characters of punctuation and AA | Partially | Optimized to tokenize English texts, but it works properly also with Italian texts |
| 2. Parsing | Syntactic analysis of the text | SPECIALIST Lexicon SPECIALIST minimal commitment parser | The SPECIALIST minimal commitment parser looks up the words of the sentence in the SPECIALIST lexicon and uses linguistic information associated to each lexical item, and constructs the syntactic representation of the text. | Yes | Strictly English-specific, it does not work properly in analyzing texts written in other languages than English |
| 3. Variant generation | Generation of variants | SPECIALIST Lexicon Lexical Variant Generation (LVG) | The algorithm generates spelling variants, abbreviations, acronyms, synonyms, inflectional and derivational variant starting from a variant generator (i.e. any sub-sequence of words that occurs in the SPECIALIST lexicon) | Yes | Strictly English-specific, it does not work properly in analyzing texts written in other languages than English |
| 4. Candidate retrieval | Identification of all the possible candidates in the Metathesaurus | "ITA-Metathesaurus" and "ENG-Metathesaurus" | The algorithm browses the UMLS Metathesaurus source to find the set of all strings containing at least one of the variants computed for the phrase | Partially, it depends on the language of the UMLS source | It depends on the language of the UMLS knowledge source, thus works properly in analyzing Italian texts |
| 5. Candidate evaluation | Evaluation of how well each candidate previously retrieved matches the text | - | The algorithm evaluates each candidate using a weighted average of four metrics: centrality, variation, coverage and cohesiveness. | No | Non language-dependent, it works properly on both English and Italian texts |
| 6. Mapping construction | Construction of the final mapping from text to the UMLS Metathesaurus | - | The algorithms combines all candidates involved in disjoint parts of the phrase, and evaluates the strength of the complete mappings. | No | Non language-dependent, it works properly on both English and Italian texts |

**2.2. Experimental setup**

Fig. 1 shows the workflow of experiments EXP1, EXP2 and EXP3.

EXP1 was aimed to provide an answer to Q1, i.e. to assess if concepts (belonging to the "Disorders" semantic group) identified in Italian clinical notes are covered in Italian knowledge sources.

In EXP1 we identified medical concepts in a set of unstructured clinical notes originally written in Italian ("ITA-TXT") and, for each identified concept, we investigated if it was possible to find a suitable description in the "ITA-Metathesaurus", which is a customized UMLS-like knowledge source containing only Metathesaurus sources in Italian (see the detailed description in section 2.5). As a comparison, we identified medical concepts also in an fully automatic translation of the original clinical notes from Italian to English ("ENG-TXT"), and we mapped them using the "ENG-Metathesaurus", which is a customized UMLS-like knowledge source containing only Metathesaurus sources in English (see the detailed description in section 2.5).

EXP2 and EXP3 were aimed to provide an answer to Q2, i.e. to understand if it is feasible to use MetaMap "as it is" to identify medical concepts in Italian clinical notes. Manual and automatic annotations with MetaMap were performed on the "ITA-TXT" and "ENG-TXT" datasets. Automated annotation was obtained using the "ITA-Metathesaurus" and the "ENG-Metathesaurus" knowledge sources. Recall, precision and F-measure were computed by comparing MetaMap's and manual annotations. Cases in which MetaMap's and manual annotations gave different results were considered as failures. A further analysis was performed to investigate different types of failures and the reason for which they occurred.

**2.3 Data corpus and normalization (EXP1, EXP2, EXP3)**

Our data corpus consists of 3462 unstructured sentences taken from the narrative portion of 100 full clinical notes written in Italian language. The original Italian clinical notes were collected during the ambulatory clinical practice at the "Policlinico" hospital in Milan (Italy) from five clinical domains, i.e. cardiology, diabetology, hepatology, nephrology, and oncology. Clinical notes have been manually anonymized by removing all personal, geographical and temporal information that could identify the patients or the clinicians. We manually normalized the text removing acronyms, abbreviation and spelling errors, obtaining the dataset we called "ITA-TXT".

**2.4 ITA to ENG translation (EXP3)**

After normalization, the "ITA-TXT" dataset was translated to English using the fully automatic translation provided by Google Translator, obtaining the "ENG-TXT" dataset. The 3462 unstructured sentences were submitted to Google Translator using a random order. We chose to use Google Translator as it is a fast and cheap translation tool.

**2.5 Knowledge sources (EXP1, EXP2, EXP3)**

Among the different knowledge sources to annotate biomedical text, the UMLS is one of the most widely used as it is constantly growing and incorporates several biomedical and health related vocabularies, in different languages [33]. Among the different UMLS sources, the Metathesaurus is a large concept-oriented database that groups terms from over 100 biomedical vocabularies into unique concepts and categorizes concepts into semantic types. The Metathesaurus contains a huge number of concepts, each identified by a Concept Unique Identifier (CUI) and organized into semantic categories [34].

In NLP processing of either biomedical literature or clinical notes, the Metathesaurus can be customized to obtain *ad hoc* Metathesaurus subsets, e.g. selecting knowledge sources in specific languages or filtering concepts using specific semantic categories [16,35]. In our experiments, we used two different knowledge sources, "ITA-Metathesaurus" and "ENG- Metathesaurus", to map the medical concepts identified in the original clinical Italian notes and their translation in English, respectively. Both "ITA-Metathesaurus" and "ENG- Metathesaurus" were based on the UMLS 2014 AA Metathesaurus. The "ITA-Metathesaurus" knowledge source was created using the MetaMap Data File Builder [36], a specific tool provided by NLM; this knowledge source contained the Italian sources of the UMLS Metathesaurus 2014 AA (details are reported in Table 2). The "ENG- Metathesaurus" knowledge source was one of the three UMLS default knowledge sources provided by MetaMap developers. In particular, the "ENG-Metathesaurus" corresponded to the wider knowledge source among those provided, the "NLM" one, and included the whole Metathesaurus (details are reported in Table 2). The "ENG-Metathesaurus" included only the English version of all the Metathesaurus sources.

Table 2. The "ITA- Metathesaurus" and "ENG- Metathesaurus" knowledge sources.

|  | "ITA- Metathesaurus" | "ENG- Metathesaurus" |
|---|---|---|
| Metathesaurus release | 2014 AA | 2014 AA |
| Metathesaurus Sources | *MedDRA* (Medical Dictionary for Regulatory Activities Terminology) *MeSH* (Medical Subject Headings) Metathesaurus Version of Minimal Standard Terminology Digestive Endoscopy International Classification of Primary Care. | Full Metathesaurus. |
| Language | Italian | English |

## 2.6 Manual annotation (EXP1, EXP2, EXP3)

Two expert annotators manually identified the medical concepts in the "ITA-TXT" and "ENG-TXT" datasets and assigned the corresponding CUIs found in the Metathesaurus knowledge sources "ITA-Metathesaurus" and "ENG-Metathesaurus". The inter-annotator agreement, calculated as percentage agreement, i.e. the percentage of concepts for which both annotators assigned the same CUI, was found to be equal to 0.75. In case of inter-annotator disagreement, a clinical adjudicator external to the annotation process decided which, between the two CUIs assigned by the two annotators, was the most proper one. Each annotated noun phrase in the "ITA-TXT" dataset and its corresponding translation in the "ENG-TXT" dataset were identified using the same CUI. When Italian medical concepts were translated incorrectly or inaccurately to English, we still mapped them to the same CUIs we used for the corresponding noun phrases in the "ITA-TXT" dataset. Normalization (see 2.3) and manual annotation were carried out simultaneously; the whole process lasted three months.

We chose to extract only medical terms belonging to the UMLS semantic categories of "Disorders" as defined by Bodenreider [34]. This choice was made for two reasons: (i) the "Disorders" semantic group is the most frequently group used in clinical notes [35]; (ii) also, it contains semantic categories (e.g. Disease or Syndrome, Symptoms) that are well represented in both English and Italian Metathesaurus sources [37]. Our choice will thus prevent to underestimate MetaMap's performances as a consequence of the lack of coverage of Italian medical terminologies. Following [34], the "Disorders" groups consisted of the following UMLS semantic types: *pathologic functions, disease or syndrome, mental or behavioral dysfunction, cell or molecular dysfunction, congenital abnormality, acquired abnormality, injury or poisoning, anatomic abnormality, neoplastic process, and virus/bacterium, sign or symptom and finding*.

## 2.7 MetaMap annotation (EXP2, EXP3)

We used MetaMap for analyzing both the "ITA-TXT" and "ENG-TXT" datasets using the two language specific "ITA-Metathesaurus" and "ENG-Metathesaurus" knowledge sources. For both cases, in addition to the default set of processing options [20], we used the option "--ignore_word_order". This option influences both the "candidate retrieval" and the "candidate evaluation" processing steps (see Table 1), allowing identifying those compound concepts in which words were not in the standard order.

Finally, as in this study we were interested in identifying all the potential medical concepts in the clinical notes, we used, for both the "ITA-Metathesaurus" and "ENG- Metathesaurus" knowledge sources the "relaxed" filtering mode [16], which allows using MetaMap in a less constrained mode, casting a wider net and giving more emphasis on recall than precision.

**2.8 Data analysis**

*2.8.1 EXP1*

We investigated how many concepts identified in the "ITA-TXT" dataset had a proper mapping in the "ITA-Metathesaurus" knowledge source. We compared results thus obtained on Italian texts with the number of concepts identified in an English translated version of the dataset, i.e. "ENG-TXT", for which we found a proper mapping in the "ENG-Metathesaurus" knowledge source.

*2.8.2 EXP2*

For each noun phrase identified by manual annotation, we analyzed the output of the MetaMap's annotation against three possible conditions: (i) the noun phrase was annotated using a correct parsing; (ii) the noun phrase was mapped to a correct CUI; (iii) in case of uncorrected parsing, (i.e. when the condition (i) has not been satisfied), we investigated whether the noun phrase has been parsed into multiple incomplete terms or if it has not been identified at all. Only exact CUI matches between manual and automatic annotation, i.e. only cases in which both the above conditions (i) and (ii) were satisfied, were considered as exact matches. All the cases different from exact matches, i.e. those corresponding to a partial mapping, mapping to a CUI different from the one identified during the manual annotation, or a missed identification of the noun phrase, were considered as failures.

Exact matches were further analyzed to count (a) how many of them were related to concepts written in the text using the "preferred term" version of "ITA-Metathesaurus" and (b) how many were related to concepts written in the text differently from the "preferred term" version of "ITA-Metathesaurus". This further analysis was aimed at quantifying how the use of MetaMap improved our results compared to results that we would have if using a simple text search of the Metathesaurus preferred terms.

We computed recall, precision, and F-measures. Recall measures the proportion of the target items that the system selected:

$$R = \frac{(\#\ exact\ matches)}{(\#\ total\ annotated\ concepts)}, \qquad (1)$$

where *#exact matches* is the number of noun phrases that satisfied both conditions (i) and (ii) and *#total annotated concepts* is the total number of concepts manually annotated in the text (considering only those noun phrases for which a suitable concept was found in the Metathesaurus).

Precision measures the proportion of selected items that the system got right:

$$P = \frac{(\# \text{ exact matches})}{(\# \text{ total concepts mapped by MetaMap})}. \quad (2)$$

where *#exact matches* is the number of concepts that satisfied both conditions (i) and (ii) and *#total concept mapped by MetaMap* is the total number of concepts that satisfied condition (i).

F-measure is the harmonic mean of precision and recall:

$$F = \frac{1}{\frac{1}{P}\alpha + (1-\alpha)\frac{1}{R}} \quad (3)$$

where P is the precision, R is the recall and the constant α determines the relative weights of precision and recall. For our task, precision and recall are equally important; we thus set α to 0.5 to give the same weigh to precision and recall.

We manually reviewed MetaMap's failures and characterize them according to three categories, as suggested by Park [22]: (1) boundary failures, (2) word sense ambiguity failures, and (3) missed term failures. A "boundary" failure occurred when a single coherent noun phrase has been parsed into multiple incomplete terms, i.e. when the condition (iii) was satisfied. A "word sense ambiguity" failure occurred if a concept has been parsed correctly but the assigned CUI was different from that resulting from the manual annotation (i.e. when condition (i) was satisfied, but condition (ii) was not satisfied). Finally, a "missed term" failure occurred when a concept has not been identified at all (i.e. no one of the three conditions was satisfied). For each category we investigated the reasons of the failures.

*2.8.3 EXP3*

Most of the data analysis implemented in EXP3 is identical to that described above for EXP2. Differently from EXP2, in EXP3 we did not perform the analysis on how many exact matches are related to concepts written in the text differently from their corresponding preferred term reported in the UMLS Metathesaurus.

**3. Results**

**3.1 EXP1**

EXP1 was aimed to provide an answer to Q1, i.e. to understand if it is possible to properly map medical concepts identified in the Italian clinical notes using the Italian knowledge sources.

Manual identification of medical concepts in the "ITA-TXT" dataset led to a total of 2077 identified concepts. We found that 58% of the identified concepts were unique concepts. Among the unique concepts, 39% were single noun phrases, while the remaining 61% were compound noun phrases. As to the normalization, only very few concepts (89, corresponding to about 4% of the whole set of identified concepts) were normalized due to misspellings or abbreviations. As shown in Table 3, 1888 concepts (about 91%) were found in the "ITA-Metathesaurus" knowledge source, while 188 concepts (about 9%) did not have a suitable concept in the Italian version of UMLS Metathesaurus.

As a comparison, we investigated if the same concepts, after a fully automatic translation from Italian to English, could be mapped properly in the "ENG-Metathesaurus" knowledge source. As shown in Table 3, 2056 concepts (about 99%) were found in the "ENG-Metathesaurus" knowledge source, while 21 concepts (about 1%) did not have a suitable concept in the English version of UMLS Metathesaurus. Also, we observed that the coverage of the "ITA-Metathesaurus" knowledge source was similar among the five clinical domains, ranging from 89.9% to 92.3%, while the coverage of the "ENG-Metathesaurus" knowledge source was higher and ranged from 98.5% to 100%. For both the "ITA-Metathesaurus" and "ENG-Metathesaurus" knowledge sources, the coverage of the Nephrology domain was the highest among all the domains. Details of the number of annotated concepts and their Italian and English Metathesaurus coverage for each specific clinical domain are reported in Table 3.

Table 3. Number of concepts (and %) identified in the "ITA-TXT" and "ENG-TXT" datasets found in the "ITA-Metathesaurus" and "ENG-Metathesaurus" knowledge sources.

|  | #annotated noun phrases | # concepts identified in the Metathesaurus | |
|---|---|---|---|
|  |  | ITA-Metathesaurus | ENG-Metathesaurus |
| cardiology | 422 | 382 (90.6%) | 416 (98.6%) |
| diabetology | 488 | 439 (90.0%) | 481 (98.6%) |
| hepatology | 402 | 367 (91.3%) | 396 (98.5%) |
| nephrology | 517 | 477 (92.3%) | 517 (100%) |
| oncology | 248 | 223 (89.9%) | 246 (99.2%) |
| total | 2077 | 1888 (90.9%) | 2056 (99.0%) |

## 3.2 EXP2

EXP2 was aimed to understand (i) how the different processing steps of by MetaMap, which are English dependent, could be used to map Italian texts; and (ii) if it is feasible to run MetaMap to map the "ITA-TXT" dataset using the "ITA- Metathesaurus" knowledge source.

As reported in Table 1, three of the six processing steps of MetaMap, i.e. tokenization, candidate evaluation and mapping construction, do not rely for their processing on English linguistic resources, thus they work properly when dealing with texts written in Italian. The candidate retrieval processing step is dependent on the language of the UMLS source, thus it works properly with texts written in Italian if the Italian UMLS knowledge source (i.e. the "ITA-Metathesaurus") is used. The remaining two steps, i.e. parsing and variant generation, are strictly built on English language, thus they do not work properly when dealing with texts written in Italian. Without using the parsing and variant generation steps, MetaMap does not generate variants for the Italian texts. We observed that, among the exact matches, there were two different types of concepts: (a) concepts written in the clinical notes using the same "preferred term" version displayed in the "ITA-Metathesaurus" knowledge source; (b) concepts written in the clinical notes differently from the "preferred term" version of "ITA-Metathesaurus" knowledge source.

As reported in Table 4, we found that about 85% of the exact matches were written in the text using the "preferred term" of the "ITA-Metathesaurus" knowledge source. The remaining 15 % were written in a way different from the "ITA-Metathesaurus" preferred term version. These concepts were written using a different word order than that used in the preferred term version. This last evidence seems to indicate that, thanks to the "ignore-word-order" processing option, MetaMap was able to identify also those concepts that were written differently from the Metathesaurus preferred term version.

We observed that the percentages of exact matches corresponding to concepts written using different word order from the preferred term version reported in the "ITA-Metathesaurus" knowledge source were almost similar among the five clinical domains, ranging from 12% to 19%.

Table 4. Medical concepts correctly identified by MetaMap in the "ITA-TXT" dataset.

|  | All the domains | Cardiology | Diabetology | Hepatology | Nephrology | Oncology |
|---|---|---|---|---|---|---|
| #exact matches | 1107 | 207 | 305 | 184 | 293 | 118 |
| #concepts written using the "preferred terms" version of "ITA-Metathesaurus" | 940 (85%) | 175 (85%) | 262 (86%) | 159 (86%) | 240 (81%) | 104 (88%) |
| #concepts written differently from the "preferred terms" version of "ITA-Metathesaurus" | 167 (15%) | 34 (15%) | 43 (14%) | 25 (14%) | 53 (19%) | 14 (12%) |

Fig.2 shows the recall (Fig.2a), precision (Fig.2b) and F-measure (Fig.2c) for the original "ITA-TXT" dataset. Mean recall calculated on the whole dataset (i.e. for all the clinical notes and all the clinical domains), was equal to 0.53. Considering the clinical domains one by one, we observed that the Nephrology and Diabetology domains showed higher recall values compared to the other clinical domains. The precision (Fig.2b) of MetaMap's annotation of the "ITA-TXT" dataset was in the range 0.97-0.99 (mean value 0.98); F-measure (Fig. 3c) was in the range 0.63-0.78 (mean value 0.69); similarly to what observed for recall, the clinical domains of Nephrology and Diabetology showed the highest F-measures values.

To analyze the effect of the normalization of misspellings and acronyms on our results, we investigated MetaMap's performances on the original Italian text without any normalization. We obtained that the mean recall, precision and F-measure were equal to 0.51, 0.95 and 0.66, respectively. Table 7 shows the comparison between results obtained without and with normalization: mean recall, precision and F-measures without the normalization were slightly lower (about 2 percentage points) than the results we obtained with the normalization.

Fig. 3a shows the results of the analysis of MetaMap's failures in the annotation of the "ITA-TXT" dataset. The "missed term", the "boundary", and the "word sense ambiguity" failures accounted for about 59%, 39%, and 2% of all the failures, respectively. Table 5 shows examples of the three types of failure. The noun phrase "edemi" (edema, in English), is an example of "missed term" failure ($1^{st}$ row); this is the plural form of the "ITA-Metathesaurus" concept "edema" (edema, in English), was not identified by MetaMap. As an example of "boundary" failure ($2^{nd}$ row), the noun phrase "stenosi vascolari" (vascular stenosis, in English), which is the plural form of the "ITA-Metathesaurus" concept "stenosi vascolare" (vascular stenosis, in English), was partially mapped by MetaMap to "stenosi" (stenosis). As an example of "word sense ambiguity" failure ($2^{rd}$ row), the noun phrase "vertigini" (dizziness, in English) was not correctly mapped to its singular form "vertigine" (dizziness, in English), but to "vertigini" (lightheadedness, in English).

Fig. 4a summarizes the reasons for the MetaMap's failures in the annotation of the "ITA-TXT" dataset. The "no variants generation", consisting in failures occurred because MetaMap did not generate variants, was the reason for which 86% of the total MetaMap's failures occurred. The remaining 14% of failures occurred for reasons different from this one and were identified as "other reasons". In particular (fig. 4a), we observed that "boundary" and "missed term" failures were mainly due to the "no variants generation" reason, while "word sense ambiguity" failures were mainly due to "other reasons".

Table 5. Examples of the different types of MetaMap's failures in the annotation of the "ITA-TXT" dataset. For each example, the table shows the noun phrase, the corresponding UMLS concept (and CUI) identified with manual annotation, the MetaMap's annotation (and, under brackets, its English translation), and the type of failure.

| Noun phrase | Manual annotation | MetaMap's annotation | Type of failure |
| --- | --- | --- | --- |
| edemi (*edema*) | C0013604: Edema (*edema*) | - | Missed term |
| stenosi vascolari (*vascular stenosis*) | C0679403: stenosi vascolare (*vascular stenosis*) | C0947637: Stenosi (*stenosis*) | Boundary |
| vertigini (*dizziness*) | C0042571: Vertigine (*dizziness*) | C0220870: Vertigini (*lightheadedness*) | Word sense ambiguity |

### 3.3 EXP3

EXP 3 aimed at investigating the performance of MetaMap to analyze an English translated version of the Italian text.

Fig.2 shows the recall (Fig.2a), precision (Fig.2b) and F-measure (Fig.2c) for the "ENG-TXT" dataset, for the whole database and the five clinical domains separately. Mean recall calculated on the whole database (i.e. for all the clinical notes and all the clinical domains), was equal to 0.75. Considering the clinical domains one by one, we observed that the Nephrology and Diabetology specialties showed higher recall values than the other clinical domains.

The precision (Fig.2b) of MetaMap's annotation of the "ENG-TXT" dataset was higher than 0.9 for all the domains; F-measure (Fig. 2c) was in the range of 0.8-0.89 (mean value 0.83); similarly to what observed for recall, the clinical domains of Nephrology and Diabetology showed higher F-measures values compared to the other clinical domains.

To analyze the effect of the normalization of misspellings and acronyms on our results, we investigated MetaMap's performances on the English translation of the original Italian text without any normalization. The mean recall, precision and F-measure were equal to 0.73, 0.9 and 0.8, respectively. Table 7 shows the comparison between results obtained without and with normalization: mean recall, precision and F-measures without the normalization were slightly lower (about 2 percentage points) than the results we obtained with the normalization.

Fig. 3b shows results of the analysis of MetaMap's failures in the annotation of the "ENG-TXT" dataset. The "boundary", the "word sense ambiguity", and the "missed term" failures accounted for 60%, 24%, and 16% of all the failures, respectively. Table 6 shows examples of the three types of failure. "Chronic pulmonary heart", an example of the "boundary" failure (1$^{st}$ row), was parsed by MetaMap into three separated terms (i.e. chronic, pulmonary and heart), while the right mapping was "chronic pulmonary heart disease". The concept "dependent edema", an example of the "word sense ambiguity" failure (2nd row), was badly mapped by MetaMap to "peripheral edema" instead of "dependent edema", because of an improper English translation. The concept "stasis", an example of the "missed term" failure (3rd row), was not mapped by MetaMap because of a bad translation from Italian to English.

Fig. 4b summarizes the reasons for which MetaMap's failures occurred in the annotation of the "ENG-TXT" dataset. The "bad ITA to ENG translation" reason, which contains all the failures due to a bad automatic translation from Italian to English, was the main cause of failure in the annotation of the "ENG-TXT" dataset (62% of all the total failures), followed by "use of medical slang" reason, which contains all the failures due to the use of specific medical slang (21% of all the total failures). Only 17% of all the failures occurred for "other reasons", which corresponds to those failures that occurred for miscellanea of reasons different from "use of medical slang" and "bad ITA to ENG translation".

Considering in details the "boundary" failures, we found that about half of them were due to "bad ITA to ENG translation", about a third of them to the "use of medical slang", and the remaining to "other reasons". For the "word sense ambiguity" failures and the "missed term" failures, we observed that they were due in a great part to "bad ITA to ENG translation" reason and in a smaller part to "use of medical slang" reason.

Table 6. Examples of the different types of MetaMap's failures in the annotation of the "ENG-TXT" dataset. For each example, the table shows the original Italian noun phrase (and, under brackets, its supervised English translation), the Google translation, the UMLS concept (and CUI) identified with manual annotation, the MetaMap's annotation, and the type of failure.

| Noun phrase | Google translation | Manual annotation | MetaMap's annotation | Type of failure |
|---|---|---|---|---|
| cuore polmonare cronico (*chronic pulmonary heart*) | chronic pulmonary heart | C0238074: Chronic Pulmonary Heart Disease | C0205191:CHRONIC C0024109:pulmonary C1281570:Heart | Boundary |
| edemi declivi (*dependent edema*) | peripheral edema | C0235437: Dependent edema | C0085649: Peripheral edema | Word sense ambiguity |
| stasi (*stasis*) | stagnation | C0333138: Stasis | - | Missed term |

Table 7. Mean recall, precision and F-measure obtained in EXP2 and EXP3, without and with the normalization.

| | EXP2 | | EXP3 | |
|---|---|---|---|---|
| | WITHOUT normalization | WITH normalization | WITHOUT normalization | WITH normalization |
| Recall | 0.51 | 0.53 | 0.73 | 0.75 |
| Precision | 0.95 | 0.98 | 0.9 | 0.93 |
| F-measure | 0.66 | 0.69 | 0.8 | 0.83 |

## 4. Discussion

In this study, we addressed two research questions: (Q1) are medical terms found in Italian clinical notes and related to the semantic group of "Disorders" properly mapped to the Italian medical terminological resources? (Q2) is it feasible to use MetaMap –as it is– to identify medical concepts related to the semantic group of "Disorders" from Italian clinical notes?

The problem of UMLS coverage for languages different from English is well known and largely investigated in the literature (see, e.g. [38,39]). To the best of our knowledge, this is the first study focused on the coverage of Italian UMLS on medical concepts that are actually found in clinical notes.

In EXP1, which addressed question Q1, we observed that 91% of the medical concepts identified in the "ITA-TXT" version of the dataset were found in the "ITA-Metathesaurus" knowledge source. As a comparison, when we analyzed the fully automatic English translated version of the dataset, i.e. the "ENG-TXT" dataset, we found, as expected, that the Metathesaurus coverage was slightly higher than with Italian, with 99% of the concepts being mapped in the "ENG-Metathesaurus". Interestingly, although the "ENG-TXT" dataset was generated as a fully automatic translation of the original "ITA-TXT" dataset, the coverage of concepts found in the "ENG-TXT" dataset is well in line with results typically found in the texts originally written in English (see e.g. [40]).

To answer to Q2, i.e. to test if it could be feasible to use MetaMap to identify medical concepts in Italian clinical notes, we carried out EXP2 and EXP3. In EXP2 we first performed MetaMap annotation on the "ITA-TXT" dataset using the "ITA-Metathesaurus" knowledge source, to understand the extent to which the different processing steps used by MetaMap (which are English dependent) could be used to process Italian texts.

The results showed that EXP2 was not trivial, as it revealed that, although MetaMap was developed to analyze texts written in English, we obtained good results also in analyzing texts written in Italian. In EXP2 we found recall and precision equal to 0.53 and 0.98, meaning that MetaMap correctly identified about half of medical concepts in the "ITA-TXT" dataset. The reason for this "not-so-bad" performance was that, as reported in Table 1, four MetaMap's processing steps work properly when analyzing texts written Italian, as three of them, i.e. tokenization, candidate evaluation and mapping construction, are not language dependent, while the remaining one, i.e. candidate retrieval, depends uniquely on the language of the UMLS source used in the analysis. Interestingly, results of EXP2 showed that about 85% of the terms correctly identified by MetaMap, were written in the text using the "preferred term" of the "ITA-Metathesaurus", while the remaining 15% were written in the texts in a different way from the "preferred term". Roughly speaking, this means that without using MetaMap, but just searching the "ITA-Metathesaurus" preferred terms in the "ITA-TXT" dataset, we would lose about 15% of the exact matches that we found in EXP2.

In EXP3 we investigated the option of using MetaMap to analyze an English translated version of the Italian text. In this experiment, the recall and precision were equal to 0.75 and 0.95, respectively. Our results are in line with those obtained in other studies which applied MetaMap to medical texts originally written in English. For example, Meystre at al. [24] used MetaMap to extract medical problems from narrative clinical notes originally written in English: in this experiment they obtained recall equal to 0.74 (which is similar to our results) and precision equal to 0.74 (which is lower than our results). Similarly to our study, Meystre et al. [24] applied a pre-processing step of "disambiguation", in which they replaced ambiguous acronyms with their full name. In another study that investigated the feasibility of using MetaMap for the retrieval of arterial branching relationships in cardiac catheterization reports written in English, recall and precision were found to be equal to 0.83 and 1, respectively [41]. Finally, a study by Pratt et al. [42] investigated MetaMap's ability to identify biomedical concepts in the titles of biomedical literature, written in English, compared to manual annotation. The authors found that MetaMap had a recall equal to 0.93 and a precision equal to 0.55, when considering both exact and partial matches [42]. We found that the recall and precision that we would have obtained without performing misspelling and acronyms normalization in the original Italian text were slightly lower (about 2 percentage points) than the results we obtained in EXP2 and EXP3. This means that although the normalization step simplifies the task of identifying medical concepts in the "ITA-TXT" and "ENG-TXT" datasets, it did not significantly affect our results.

The approach of combining fully automatic English translation and MetaMap we used in EXP3 is very similar to the one described in [31,32]. Even if the approach of combining Google Translator and MetaMap was previously used our experiment was not trivial as we had completely different experimental conditions from the Spanish researchers [32]. As a matter of fact, our experimental conditions involved a different language, - Italian in our study, Spanish in [32] – and a different type of texts - clinical notes in our study, articles from biomedical literature in [32]. Using a different language could affect both the accuracy of the Google translation and the coverage of the UMLS Metathesaurus, and thus could influence the results of the experiment. In addition, using clinical notes instead of biomedical literature could affect both the quality of the translation and the performance of MetaMap. As a matter of fact, clinical notes have peculiar grammatical structure, show variability in the formatting and extensively use medical slang. Considering these differences in the experimental conditions, it might be interesting to compare our results and those reported by the Spanish researchers. In EXP3 we obtained a recall equal to 0.75, which is slightly higher than the mean value of similarity equal to 0.71 calculated across the various experiments reported by the Spanish researchers, where the similarity reported in [31,32] is a performance measure very similar to the recall.

In EXP2 we obtained lower recall and higher precision than in EXP3. This result was not surprising, as MetaMap processing is heavily based on the variant generation processing step and, as reported in Table 1, the variant generation processing step does not work properly with texts written in languages different from English.

We identified three types of MetaMap failures: (1) boundary failures, (2) word sense ambiguity failures, and (3) missed term failures. In EXP2, i.e. when using MetaMap to annotate the Italian version of the dataset, we obtained a high number of "missed terms" failures and "boundary" failures, equal to 59% and 39% of all the failures, respectively, and a very low number of "word sense ambiguity" failures, equal to 2%. We discovered that MetaMap's failures in the "ITA-TXT" dataset were mainly due to the "no variants generation" reason, i.e. they occurred because MetaMap did not allow to generate variants for Italian noun phrases. This finding was expected, because, as discussed above, the MetaMap "variant generation" processing step (see Table 1) is based on English linguistic resources and thus does not work to generate variants in languages different from English. A possible strategy that could be used to overcome this issue could be to adapt MetaMap to generate variants for Italian noun phrases.

In EXP3, i.e. when using MetaMap to annotate the fully automatic English translation of the dataset, the prevalence of the three types of failures was equal to 60%, 24% and 16% for "boundary" failures, "word sense ambiguity" failures and "missed term" failures, respectively. Our results were different from those reported in other studies: for example, Park et al.[22], when investigating MetaMap's failures in annotating patient-generated text, found that 82.2% were "word sense ambiguity" failures, 15.9% were "boundary" failures and 1.9% were "missed term" failures. In another study [40] it was found that "boundary" failures, "word sense ambiguity" failures were, respectively, 5% and 4% of all the failures in annotating disease summary documents using MetaMap. These differences could be due to the specific approach we used in the analysis: instead of using MetaMap to annotate texts originally written in English, we used MetaMap to annotate clinical notes translated from Italian to English using a fully automated translation tool.

We also found that 62% of all the failures were due to a "bad ITA to ENG translation". This finding might suggest that a more accurate translation of medical terms could be of some help to reduce the number of MetaMap's failures. The second main reason of failure, which caused 21% of all the failures, was found to be the "use of medical slang". The presence of specific "medical slang" is a well-known characteristics of clinical notes [12]. Annotating texts containing "medical slang" is a challenging task as "medical slang" is not codified in the UMLS. Moreover, creating a specific dictionary to code "medical slang" could be difficult, as "medical slang" could be very specific to the clinical context in which it has been used. Maybe, machine-learning techniques could be used to create ad hoc dictionaries containing most recurrent "medical slang" for a specific clinical domain.

## 5. Conclusions

We found that most of the medical terms found in Italian clinical notes could be properly mapped using the Italian UMLS Metathesaurus sources. We observed that four of six MetaMap's processing steps worked properly in analyzing the original Italian dataset "ITA-TXT"; this allowed to correctly identify about half of the concepts. Interestingly, using MetaMap to annotate the "ITA-TXT" dataset instead of using a simple text search improved our results of about 15 percentage points. When using MetaMap to annotate the original Italian dataset, we obtained poorer performances than those described in the literature for the annotation of English written texts, with lower recall and F-measures, but higher precision. Moreover, we found that most of the failures occurred because MetaMap did not allow generating meaningful variants for the Italian language. This suggests that modifying MetaMap with a proper "Italian SPECIALIST lexicon" to allow generating meaningful Italian variants could improve the performance.

On the contrary, MetaMap's performances in mapping the English translated dataset "ENG-TXT" were in line with findings in literature, with comparable recall, and even higher precision. We observed that most of the failures were due to a bad Italian to English translation, therefore using an automatic translation tool specialized in translating medical concepts might allow to reach better performances.

In conclusion, the performances obtained on the fully automatic translation of the Italian text are good enough to allow to use MetaMap "as it is" in clinical practice. The next step in our work will involve the evaluation of the performance of the method when processing concepts belonging to other semantic groups, e.g., treatments and diagnostic procedures, and the handling of context information, such as negation, temporality and uncertainty. Moreover, it would be interesting to investigate the accuracy of the proposed method in specific research tasks (e.g. detecting drug adverse events, or investigating relationships between specific treatments and follow-up of the disease) related to specific clinical domains (e.g. audiology).

**Acknowledgements**

The activities here described come within the Scientific Agreement on "Informatica Clinica & Sanità Digitale di Dominio" (Progress in Clinical Informatics) among the 'Politecnico di Milano Dipartimento di Elettronica, Informazione e Bioingegneria', 'Fondazione IRCCS Ca' Granda Ospedale Maggiore Policlinico', and 'Istituto di Elettronica e di Ingegneria dell'Informazione e delle Telecomunicazioni del Consiglio Nazionale delle Ricerche' (2015-2017).

The authors gratefully acknowledge Dr. Demner-Fushman, Dr. Lang and Dr. Aronson from the Lister Hill Center, U.S. National Library of Medicine, for their useful help in methodological support and their comments during the design of the protocol.

**References**

[1] S.M. Meystre, G.K. Savova, K.C. Kipper-Schuler, J.F. Hurdle, Extracting information from textual documents in the electronic health record: a review of recent research, Yearb Med Inf. 35 (2008) 128–144.

[2] D. Demner-Fushman, W.W. Chapman, C.J. McDonald, What can natural language processing do for clinical decision support?, J. Biomed. Inform. 42 (2009) 760–772. doi:10.1016/j.jbi.2009.08.007.

[3] V.M. Pai, M. Rodgers, R. Conroy, J. Luo, R. Zhou, B. Seto, Workshop on using natural language processing applications for enhancing clinical decision making: an executive summary, J. Am. Med. Informatics Assoc. 21 (2014) e2–e5. doi:10.1136/amiajnl-2013-001896.

[4] S. Pradhan, N. Elhadad, B.R. South, D. Martinez, L. Christensen, A. Vogel, et al., Evaluating the state of the art in disorder recognition and normalization of the clinical narrative., J. Am. Med. Inform. Assoc. 22 (2015) 143–54. doi:10.1136/amiajnl-2013-002544.

[5] W.W. Chapman, M. Fiszman, J.N. Dowling, B.E. Chapman, T.C. Rindflesch, Identifying respiratory findings in emergency department reports for biosurveillance using MetaMap., Stud. Health Technol. Inform. 107 (2004) 487–91.

[6] L. Cui, S.S. Sahoo, S.D. Lhatoo, G. Garg, P. Rai, A. Bozorgi, et al., Complex epilepsy phenotype extraction from narrative clinical discharge summaries., J. Biomed. Inform. 51 (2014) 272–9. doi:10.1016/j.jbi.2014.06.006.

[7] C. Shivade, P. Raghavan, E. Fosler-Lussier, P.J. Embi, N. Elhadad, S.B. Johnson, et al., A review of approaches to identifying patient phenotype cohorts using electronic health records., J. Am. Med. Inform. Assoc. 21 (2013) 221–30. doi:10.1136/amiajnl-2013-001935.

[8] D. a Hanauer, M. Saeed, K. Zheng, Q. Mei, K. Shedden, A.R. Aronson, et al., Applying MetaMap to Medline for identifying novel associations in a large clinical dataset: a feasibility analysis., J. Am. Med. Inform. Assoc. (2014) 925–937. doi:10.1136/amiajnl-2014-002767.

[9] P.B. Jensen, L.J. Jensen, S. Brunak, Mining electronic health records: towards better research applications and clinical care, Nat. Rev. Genet. 13 (2012) 395–405. doi:10.1038/nrg3208.

[10] X. Wang, G. Hripcsak, M. Markatou, C. Friedman, Active Computerized Pharmacovigilance Using Natural

Language Processing, Statistics, and Electronic Health Records: A Feasibility Study, J. Am. Med. Informatics Assoc. 16 (2009) 328–337. doi:10.1197/jamia.M3028.

[11] S. V Iyer, R. Harpaz, P. LePendu, A. Bauer-Mehren, N.H. Shah, Mining clinical text for signals of adverse drug-drug interactions., J. Am. Med. Inform. Assoc. 21 (2014) 353–62. doi:10.1136/amiajnl-2013-001612.

[12] R. Leaman, R. Khare, Z. Lu, Challenges in clinical natural language processing for automated disorder normalization, J. Biomed. Inform. 57 (2015) 28–37. doi:10.1016/j.jbi.2015.07.010.

[13] D. Jurafsky, J.H. Martin, Speech and Language Processing: An introduction to Natural Language Processing, Computational Linguistics, and Speech Recognition (Introduction), 2008. doi:10.1016/S0065-230X(09)04001-9.

[14] M. Skeppstedt, M. Kvist, G.H. Nilsson, H. Dalianis, Automatic recognition of disorders , findings , pharmaceuticals and body structures from clinical text : An annotation and machine learning study, J. Biomed. Inform. 49 (2014) 148–158. doi:10.1016/j.jbi.2014.01.012.

[15] E. Castro, A. Iglesias, P. Martínez, L. Castaño, Automatic identification of biomedical concepts in spanish-language unstructured clinical texts, Proc. 1st ACM Int. Heal. Informatics Symp. (2010) 751–757. doi:10.1145/1882992.1883106.

[16] D. Demner-fushman, J.G. Mork, S.E. Shooshan, A.R. Aronson, UMLS content views appropriate for NLP processing of the biomedical literature vs . clinical text, J. Biomed. Inform. 43 (2010) 587–594. doi:10.1016/j.jbi.2010.02.005.

[17] N. Sager, M. Lyman, C. Bucknall, N. Nhan, L.J. Tick, Natural Language Processing and the Representation of Clinical Data, J. Am. Med. Informatics Assoc. 1 (1994) 142–160. doi:10.1136/jamia.1994.95236145.

[18] C. Friedman, G. Hripcsak, W. DuMouchel, S.B. Johnson, P.D. Clayton, Natural language processing in an operational clinical information system, Nat. Lang. Eng. 1 (1995) 83–108. doi:10.1017/S1351324900000061.

[19] C. Friedman, Towards a comprehensive medical language processing system: methods and issues., Proc. AMIA Annu. Fall Symp. (1997) 595–9.

[20] A. Aronson, Effective mapping of biomedical text to the UMLS Metathesaurus: the MetaMap program., in: AMIA Annu. Symp. Proc., 2001: pp. 17–21.

[21] P.F. Brennan, A.R. Aronson, Towards linking patients and clinical information: detecting UMLS concepts in e-mail, J. Biomed. Inform. 36 (2003) 334–341. doi:10.1016/j.jbi.2003.09.017.

[22] A. Park, A. Hartzler, J. Huh, D.W. McDonald, W. Pratt, Automatically detecting failures in natural language processing tools for online community text, J. Med. Internet Res. 17 (2015). doi:10.2196/jmir.4612.

[23] S. Meystre, P.J. Haug, Evaluation of Medical Problem Extraction from Electronic Clinical Documents Using MetaMap Transfer (MMTx)., Stud. Health Technol. Inform. 116 (2005) 823–828.

[24] S. Meystre, P.J. Haug, Natural language processing to extract medical problems from electronic clinical documents: Performance evaluation, J. Biomed. Inform. 39 (2006) 589–599. doi:10.1016/j.jbi.2005.11.004.

[25] P. Lependu, S. V Iyer, C. Fairon, N.H. Shah, Annotation Analysis for Testing Drug Safety Signals using Unstructured Clinical Notes., J. Biomed. Semantics. 3 Suppl 1 (2012) S5. doi:10.1186/2041-1480-3-S1-S5.

[26] O. Bodenreider, The Unified Medical Language System (UMLS): integrating biomedical terminology., Nucleic Acids Res. 32 (2004) D267–70. doi:10.1093/nar/gkh061.

[27] M. Oronoz, A. Casillas, K. Gojenola, A. Perez, Automatic Annotation of Medical Records in Spanish with Disease, Drug and Substance Names, in: Prog. Pattern Recognition, Image Anal. Comput. Vision, Appl., Springer, Berlin Heidelberg, 2013: pp. 536–543. doi:10.1007/978-3-642-41827-3_67.

[28] T. Delbecque, P. Zweigenbaum, MetaCoDe: A Lightweight {UMLS} Mapping Tool, in: Artif. Intell. Med., Springer Berlin Heidelberg, 2007: pp. 242–246.

[29] L. Deléger, C. Grouin, P. Zweigenbaum, Extracting medication information from French clinical texts, Stud. Health Technol. Inform. 160 (2010) 949–953. doi:10.3233/978-1-60750-588-4-949.

[30] L.. Ferreira, A.. Teixeira, J. Cunha, Medical information extraction in European Portuguese, 2013. doi:10.4018/978-

1-4666-3986-7.ch032.

[31] F. Carrero, J.C. Cortizo, J.M. Gomez, M. De Buenaga, In the Development of a Spanish Metamap, in: Proc. 17th ACM Conf. Inf. Knowl. Manag. ACM, 2008: pp. 1465–1466.

[32] F. Carrero, J.C. Cortizo, J.M. Gomez, Building a Spanish MMTx by using automatic translation and biomedical ontologies, in: Intell. Data Eng. Autom. Learn., 2008: pp. 346–353. doi:10.1007/978-3-540-88906-9-44.

[33] National Library of Medicine, UMLS® Reference Manual, (2009). http://www.ncbi.nlm.nih.gov/books/NBK9676/.

[34] O. Bodenreider, A.T. McCray, Exploring semantic groups through visual approaches, J. Biomed. Inform. 36 (2003) 414–432. doi:10.1016/j.jbi.2003.11.002.

[35] S.T. Wu, H. Liu, D. Li, C. Tao, M.A. Musen, C.G. Chute, et al., Unified Medical Language System term occurrences in clinical notes: a large-scale corpus analysis., J. Am. Med. Inform. Assoc. 19 (2012) e149–56. doi:10.1136/amiajnl-2011-000744.

[36] National Library of Medicine, MetaMap Data File Builder, (n.d.). https://metamap.nlm.nih.gov/DataFileBuilder.shtml.

[37] National Library of Medicine, UMLS Source Vocabulary Documentation, (n.d.). https://www.nlm.nih.gov/research/umls/sourcereleasedocs/index.html.

[38] M. Volk, B. a. Ripplinger, Š. Vintar, P. Buitelaar, D. Raileanu, B. Sacaleanu, Semantic annotation for concept-based cross-language medical information retrieval, Int. J. Med. Inform. 67 (2002) 97–112. doi:10.1016/S1386-5056(02)00058-8.

[39] K. Markó, S. Schulz, U. Hahn, MorphoSaurus--design and evaluation of an interlingua-based, cross-language document retrieval engine for the medical domain., Methods Inf. Med. 44 (2005) 537–45.

[40] G. Divita, T. Tse, L. Roth, Failure analysis of MetaMap transfer (MMTx), Stud. Health Technol. Inform. 107 (2004) 763–767. doi:10.3233/978-1-60750-949-3-763.

[41] T.C. Rindflesch, C. a Bean, C. a Sneiderman, Argument identification for arterial branching predications asserted in cardiac catheterization reports., in: AMIA Annu. Symp. Proc., 2000: pp. 704–8.

[42] W. Pratt, M. Yetisgen-Yildiz, A study of biomedical concept identification: MetaMap vs. people., in: AMIA Annu. Symp. Proc., 2003: pp. 529–33. doi:D030003464 [pii].

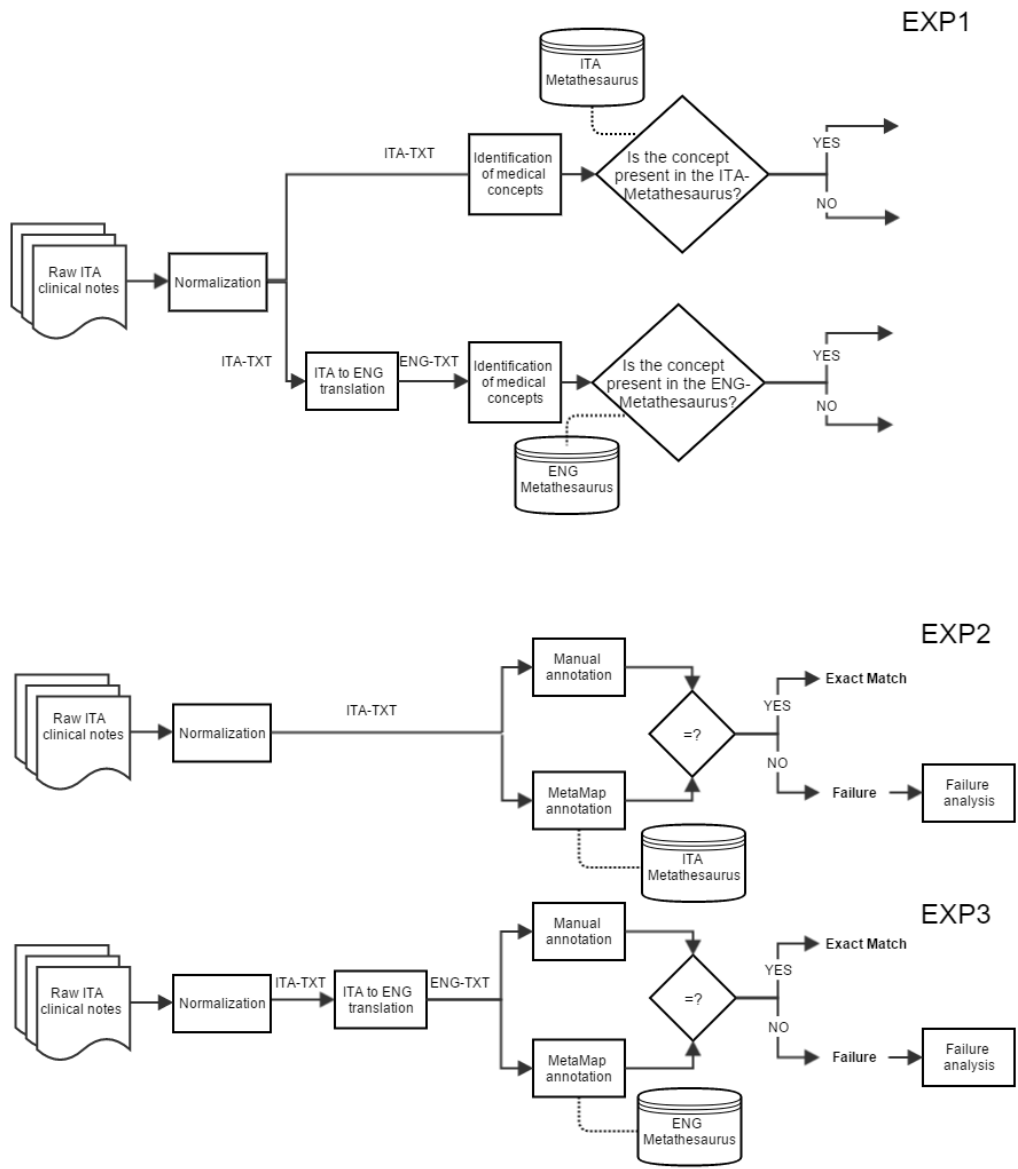

Fig. 1. Flowchart of experiments EXP1, EXP2 and EXP3.

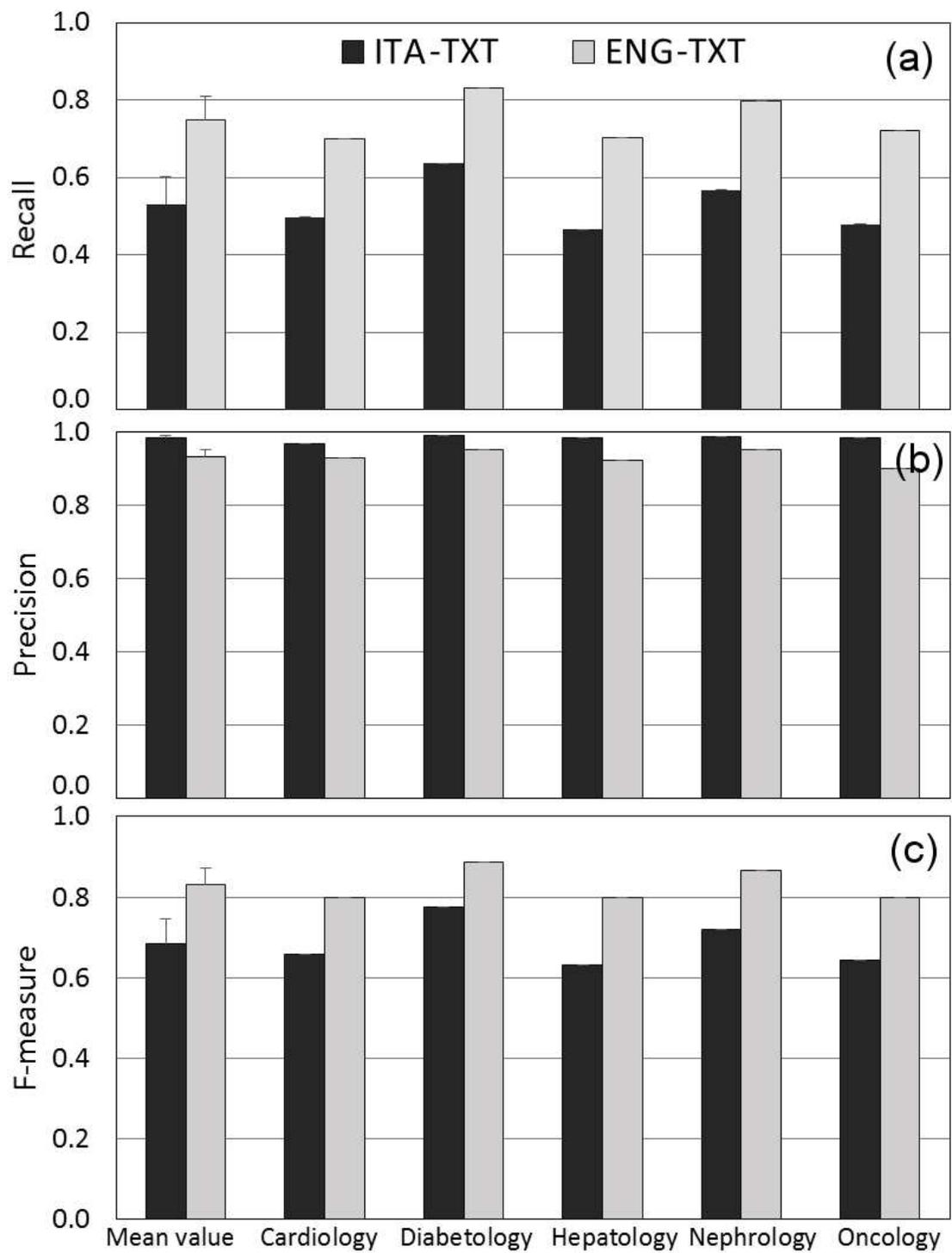

Fig. 2. Recall (3a), precision (3b) and F-measure (3c) for "ITA-TXT" and "ENG-TXT" datasets.

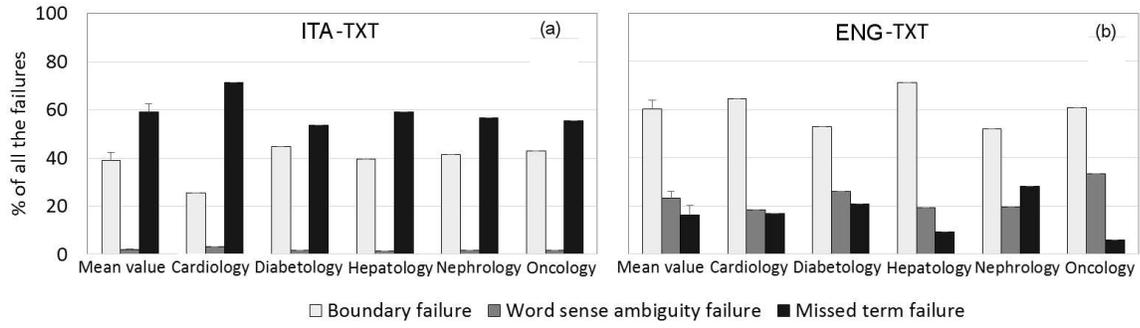

Fig. 3. Analysis of failures in the annotation of (a) "ITA-TXT" and (b) "ENG-TXT" datasets.

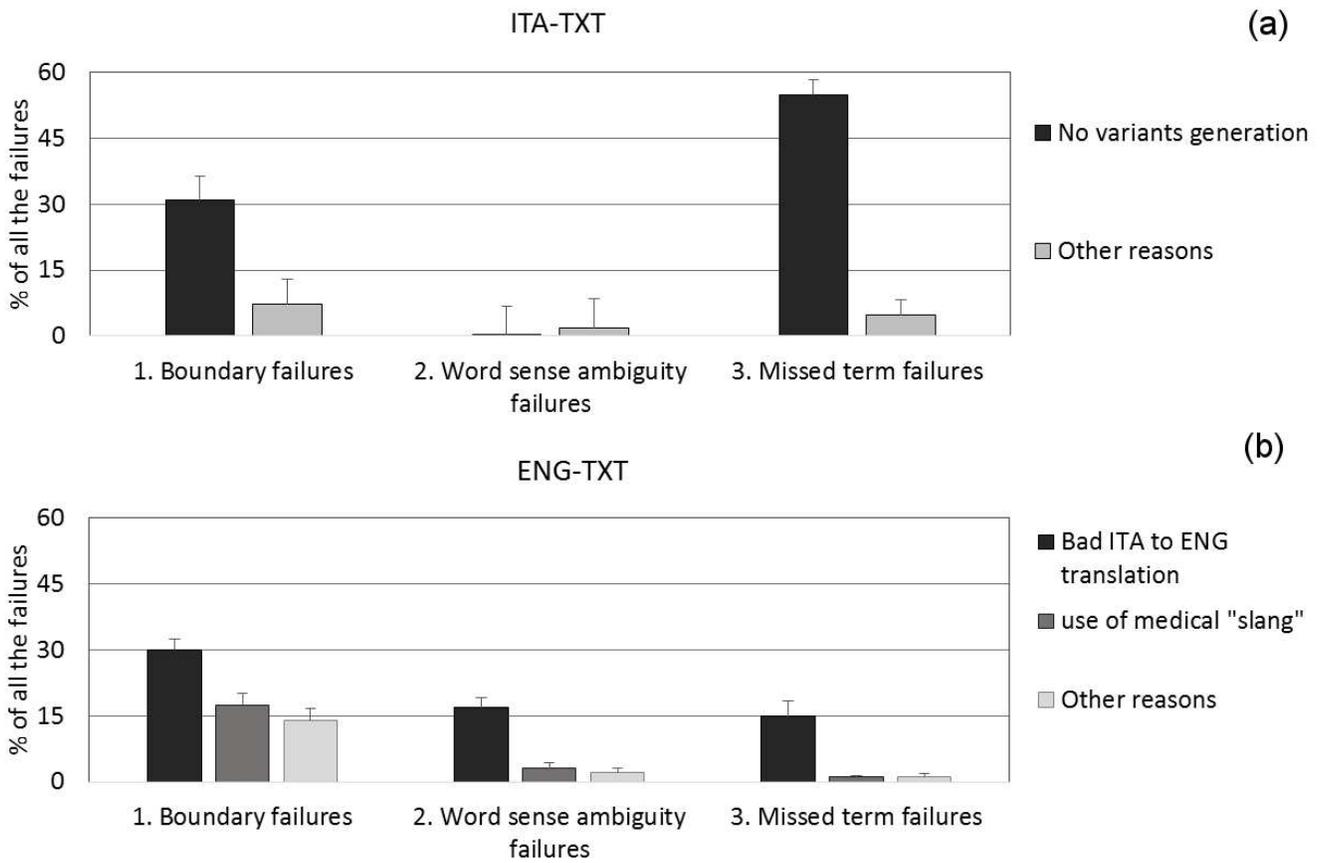

Fig. 4. Analysis of the reasons for which the different types of annotation failures occurred in (a) "ITA-TXT" and (b) "ENG-TXT" datasets.